\documentclass[10pt,twocolumn,letterpaper]{article}

\usepackage{cvpr}
\usepackage{times}
\usepackage{epsfig}
\usepackage{amsmath, amssymb, amsthm, enumerate, graphicx}
\usepackage{bm}
\usepackage{xcolor}
\usepackage{marvosym}
\usepackage{wasysym}
\usepackage{fancyhdr}
\usepackage{graphicx}
\usepackage{amsmath}
\usepackage{amssymb}
\usepackage{multirow}
\usepackage{multicol}
\usepackage{subfigure}
\usepackage{amsmath}
\usepackage{graphicx}
\usepackage{bbm}
\usepackage{amssymb}
\usepackage{symmath}

\usepackage[pagebackref=true,breaklinks=true,letterpaper=true,colorlinks,bookmarks=false]{hyperref}

\cvprfinalcopy % *** Uncomment this line for the final submission

\begin{document}

%%%%%%%%% TITLE
\title{MoBiNet: A Mobile Binary Network for Image Classification}

% Authors at the same institution
%\author{First Author \hspace{2cm} Second Author \\
% Authors at different institutions
\author{Hai Phan \\
Carnegie Mellon University\\
{\tt\small haithanp@andrew.cmu.edu}
\and
Dang Huynh \\
Axon\\
{\tt\small dhuynh@axon.com}
\and
Yihui He \\
Carnegie Mellon University\\
{\tt\small he2@andrew.cmu.edu}
\and
Marios Savvides \\
Carnegie Mellon University\\
{\tt\small marioss@andrew.cmu.edu}
\and
Zhiqiang Shen \\
Carnegie Mellon University\\
{\tt\small zhiqians@andrew.cmu.edu}
}

\maketitle
% \ifwacvfinal\thispagestyle{empty}\fi

%%%%%%%%% ABSTRACT
\begin{abstract}
   MobileNet and Binary Neural Networks are two among the most widely used techniques to construct deep learning models for performing a variety of tasks on mobile and embedded platforms.
   %The MobileNet can significantly reduce computational cost thanks to the use of separable depth-wise convolution, while Binary Neural Network is an extreme case of network quantization where one-bit parameters can optimally approximate float-type model.
   In this paper, we present a simple yet efficient scheme to exploit MobileNet binarization at activation function and model weights. However, training a binary network from scratch with separable depth-wise and point-wise convolutions in case of MobileNet is not trivial and prone to divergence. To tackle this training issue, we propose a novel neural network architecture, namely MoBiNet - Mobile Binary Network in which skip connections are manipulated to prevent information loss and vanishing gradient, thus facilitate the training process.
   %The skip connections are designed in three basic modules: Pre-block, Mid-block, and Post-block.
   More importantly, while existing binary neural networks often make use of cumbersome backbones such as Alex-Net, ResNet, VGG-16 with float-type pre-trained weights initialization, our MoBiNet focuses on binarizing the already-compressed neural networks like MobileNet without the need of a pre-trained model to start with. Therefore, our proposal results in an effectively small model while keeping the accuracy comparable to existing ones. Experiments on ImageNet dataset show the potential of the MoBiNet as it achieves \textbf{54.40}\% top-1 accuracy and dramatically reduces the computational cost with binary operators. 

\end{abstract}

%%%%%%%%% BODY TEXT
\section{Introduction}

The rising of Convolution Neural Networks (CNNs) in Deep Learning has resulted in a variety of significant improvements in complicated tasks such as object detection~\cite{Dai-NIPS2016,girshick2014rcnn,Ren-Faster,Wei-SSD,Luu_FD_FG2018,Luu_FD_CVPR2018,shen2017dsod,ZhiShen19,Zhu_2019_CVPR,he2019bounding,shen2019object}, object segmentation~\cite{he2017maskrcnn,Long-FCN,Luu_Seg_PR2018,Luu_Seg_TIP2018}, text classification~\cite{Kowsari2018RMDL,Sachan2019RevisitingLN}, etc. These impressive outcomes have been attributed to the learning capability of the million-parameter structure of convolutional layers in neural networks. %The more dataset is used to train networks, the more number of parameters of CNNs is required to fit the data points.
Intuitively saying, the complexity of CNNs should be proportional to its capacity and hence, there always exists a trade-off between accuracy and computational cost of a CNN model.

Image classification ~\cite{Kaiming_ResNet,Alex_NIPS2012,Simonyan-VGGNet} has attracted many research efforts in recent years.  ImageNet~\cite{imagenet_cvpr09} stands out to be one of the most popular datasets used for evaluating classification accuracy. This dataset contains millions of images categorized into a wide range of contexts which make the classification task extremely challenging. To cope with this, many models have been proposed: Alex-Net~\cite{Alex_NIPS2012}, VGG-16~\cite{Simonyan-VGGNet}, Inception~\cite{Szegedy2016RethinkingTI}, ResNet~\cite{Kaiming_ResNet}. These networks are powerful but cumbersome, therefore the deployment on portable and mobile devices is not preferable. To be more specific, Alex-Net and VGG-16 Caffe~\cite{Yangging-Caffe} models are over 200MB and 500MB in size, and take 725M FLOPs and 16 GFLOPs respectively in terms of computational cost. To better suit mobile devices, some techniques have emerged, shedding light on deep compressed neural networks, including model pruning~\cite{Han-LBWs,Han-Deep,Liu2017LearningEC,he2017channel}, light-weight modules~\cite{Howard-MobileNet,Sandler-MNetv2,he2019addressnet}, binary network~\cite{Itay-BNN,Matt-BinaryNet,Rastegari-xnor,CourbariauxBD15} and mimicked network~\cite{Li2017MimickingVE,Wei2018QuantizationMT}. 

Building a small but efficient neural network is not trivial. We want to optimize the number of model parameters while preserving comparable accuracy and facilitating ease of training.
Pruning methods shrink model size by eliminating insignificant channels that cause redundant computation. However, the search for such channels is expensive and requires to be done in an attentive manner. Light-weight networks such as SqueezeNet~\cite{Iandola2017SqueezeNetAA}, MobileNet~\cite{Howard-MobileNet, Sandler-MNetv2}, ShuffleNet~\cite{Zhang2018ShuffleNetAE,Ma_2018_ECCV}, ESPNet~\cite{mehta2018espnet} achieve promising results for not only image classification but object detection and segmentation as well. 

Recently applying ideas of AutoML, ~\cite{He_2018_ECCV} takes advantage of reinforcement learning to improve the network compression. Additionally, an innovative neural network called Neural Architecture Search (NASNet) ~\cite{45826,47144,pmlr-v80-pham18a} examines a database of dimension convolution layers to automate an effective architecture design, surpassing prior human-defined neural networks. However, the search for proper modules is terribly complicated and demands a vast amount of computational resources to serve an exhaustive training process. Another approach is network binarization which consists of manipulating boolean-type parameters to approximate the deep neural network calculation, so-called Binary Neural Networks (BNNs)~\cite{Matt-BinaryNet,CourbariauxBD15}. There are several advantages of BNNs. Firstly, thanks to the use of 1-bit representation, the model shrinkage is guaranteed by a proportional factor 32x compared to the traditional 32-bit float type. Secondly, with binary, multiplication of activations and 1-bit weights can be replaced by the bit-wise operation that undoubtedly results in a speed-up. Moreover, xnor and popcount operators are utilized to expedite the neural networks~\cite{Rastegari-xnor}.
From the engineering perspective, these operators can be easily implemented on either GPU or CPU-based devices.
Finally, even though the binary network's accuracy trade-off is not always negligible in comparison with the float-type counterparts, impressive reduction of computational expense and storage requirement makes deep binary neural network potential for practical applications.

\begin{figure*}
\begin{center}
\subfigure[Standard convolutional module]{
    \includegraphics[scale=0.2]{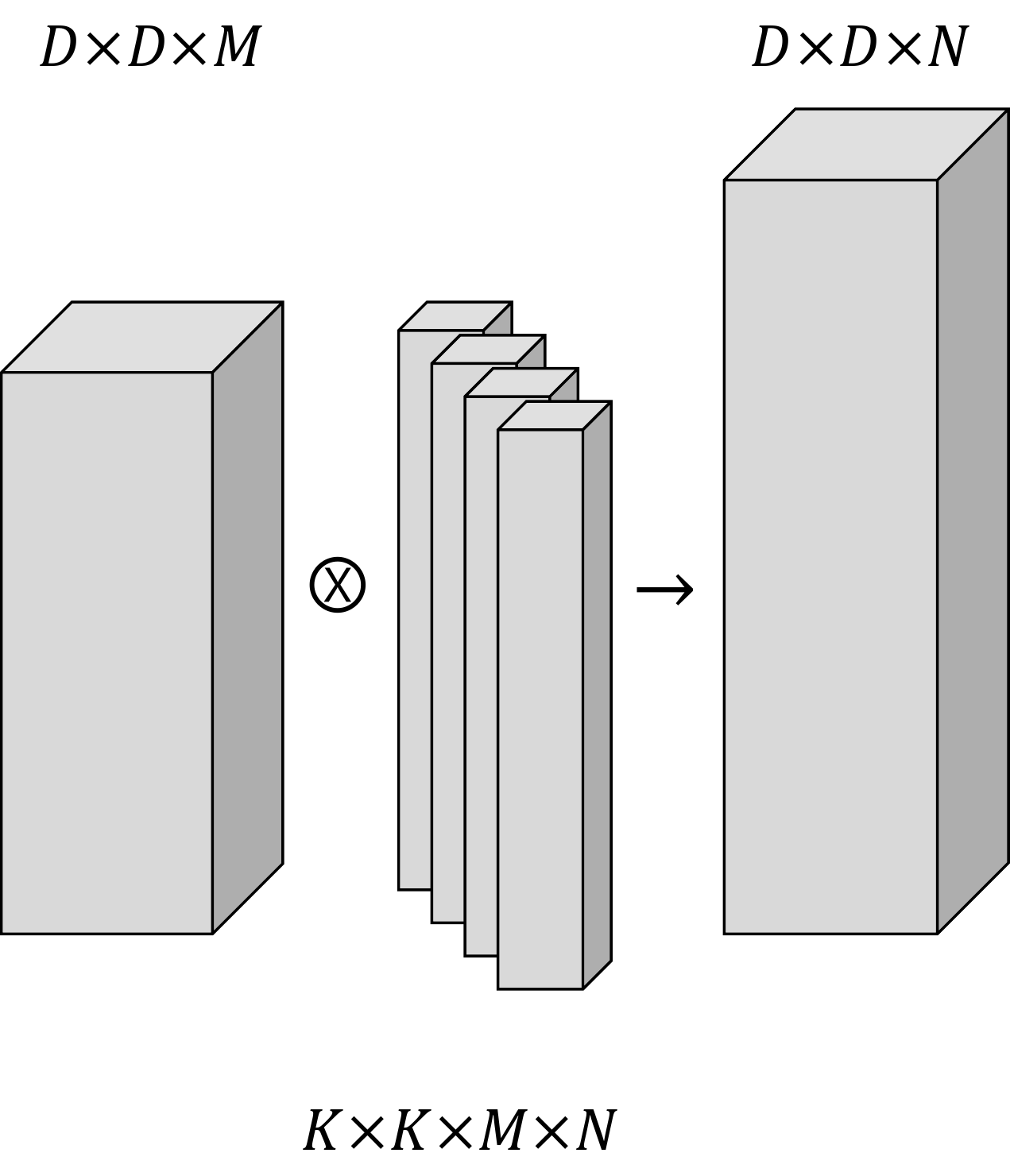}
    % \caption{demo a}
}\hspace{0.5cm}
\subfigure[MobileNet v1 dw-module~\cite{Howard-MobileNet}]{
    \includegraphics[scale=0.2]{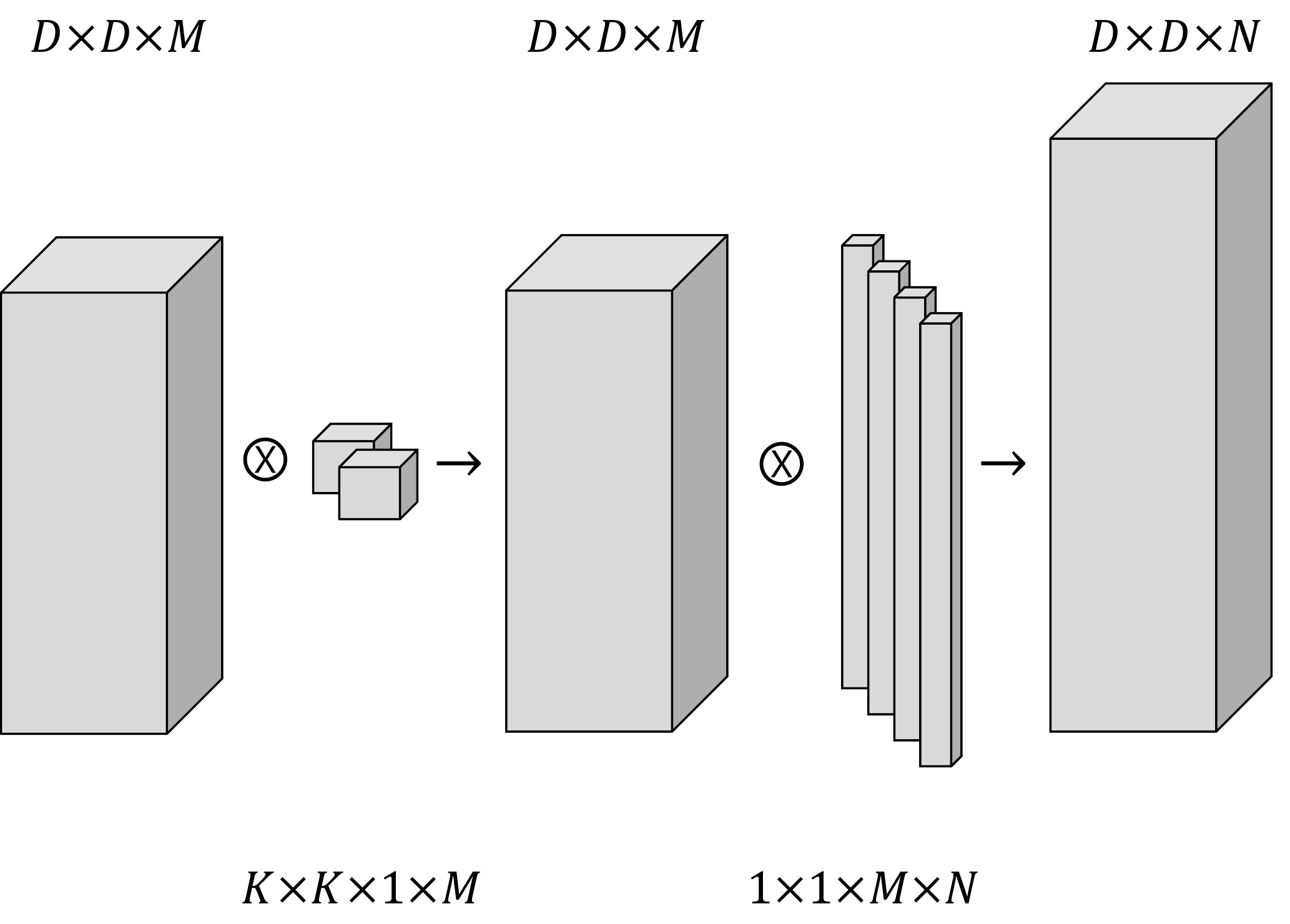}
    % \caption{demo b}
}\hspace{0.5cm}
\subfigure[MoBiNet Pre-block]{
    \includegraphics[scale=0.2]{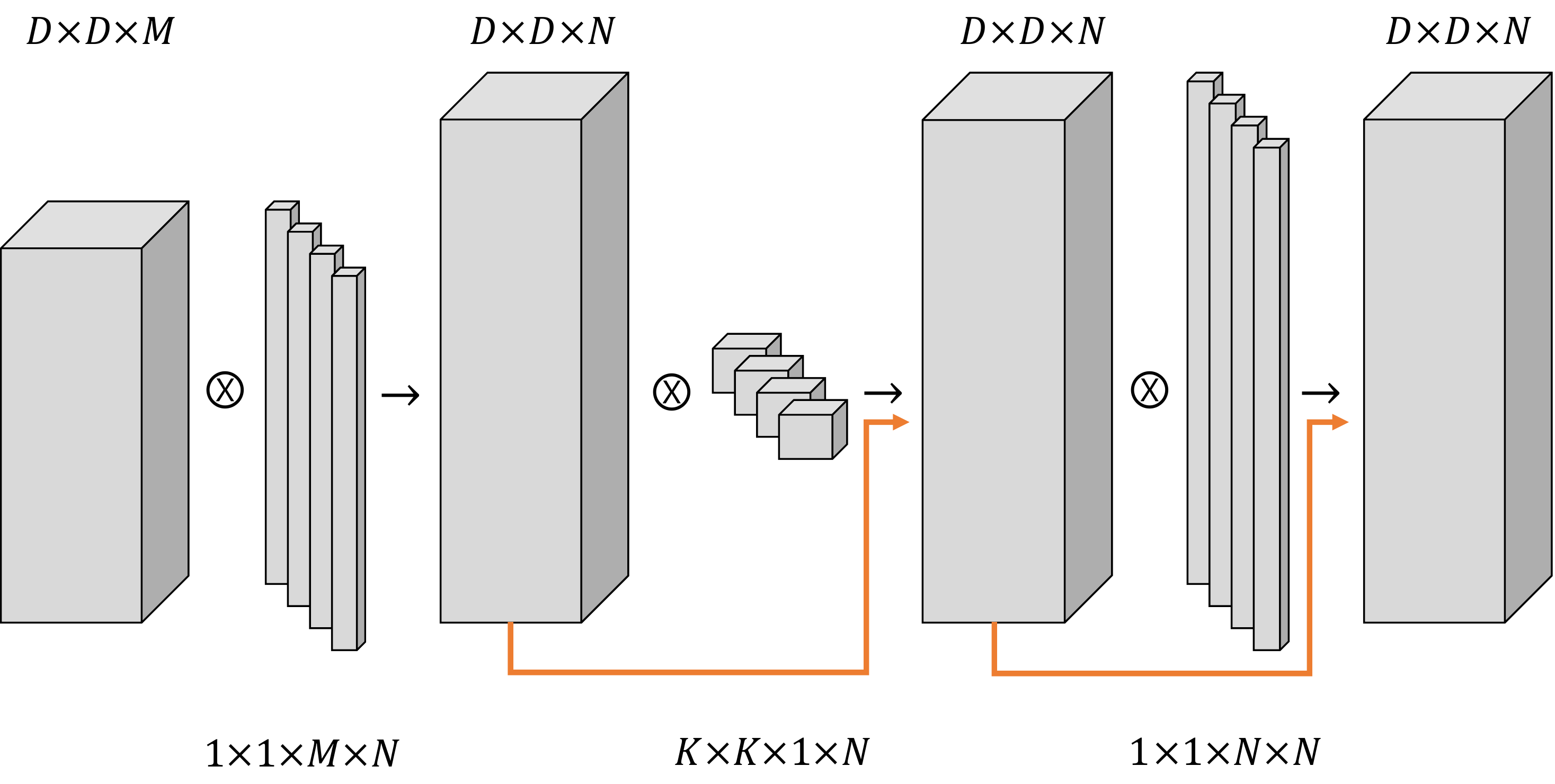}
    % \caption{demo b}
}\hspace{0.5cm}
\subfigure[MoBiNet Mid-block]{
    \includegraphics[scale=0.2]{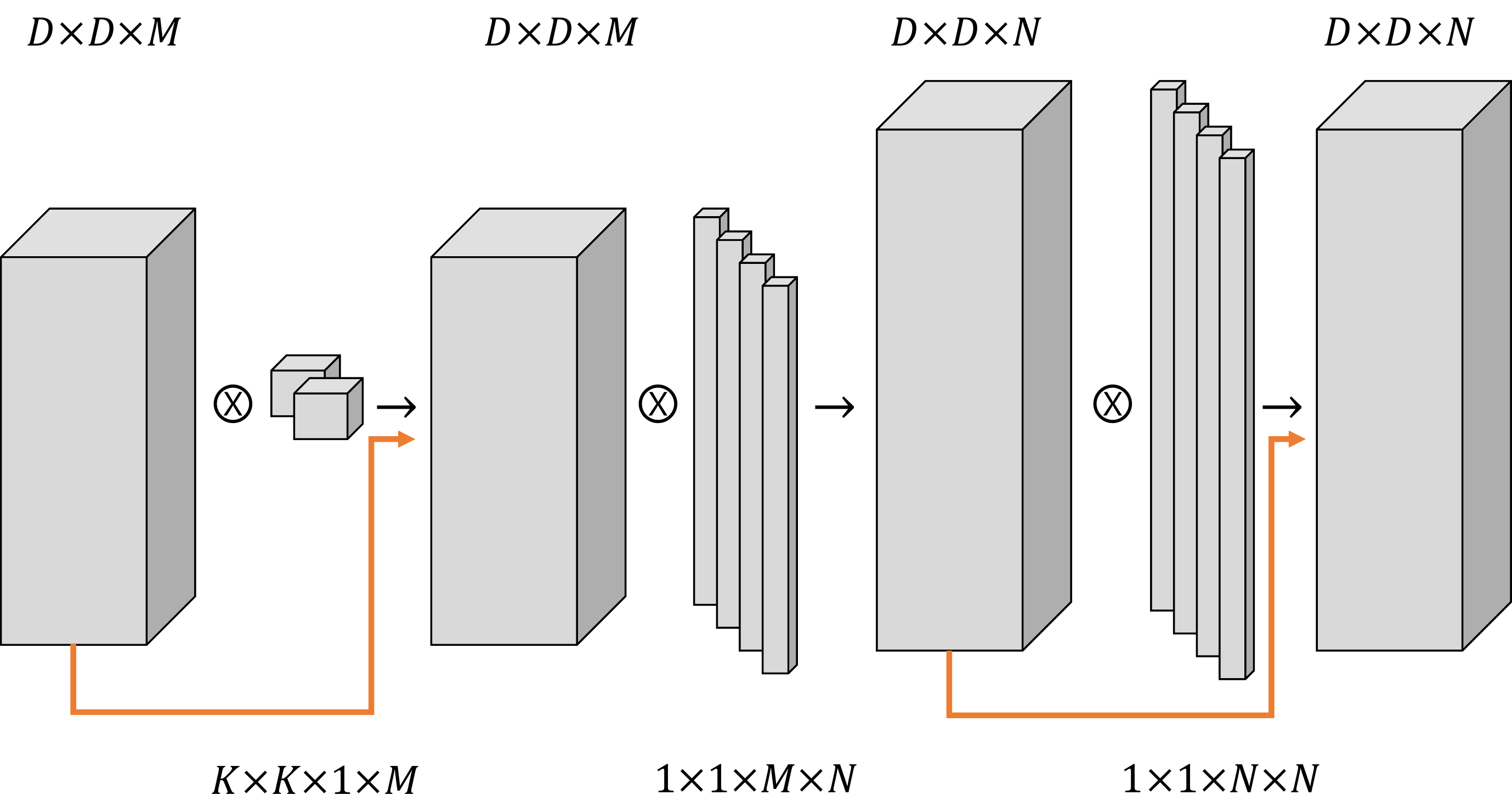}
    % \caption{demo b}
}\hspace{0.5cm}
\subfigure[MoBiNet Post-block]{
    \includegraphics[scale=0.2]{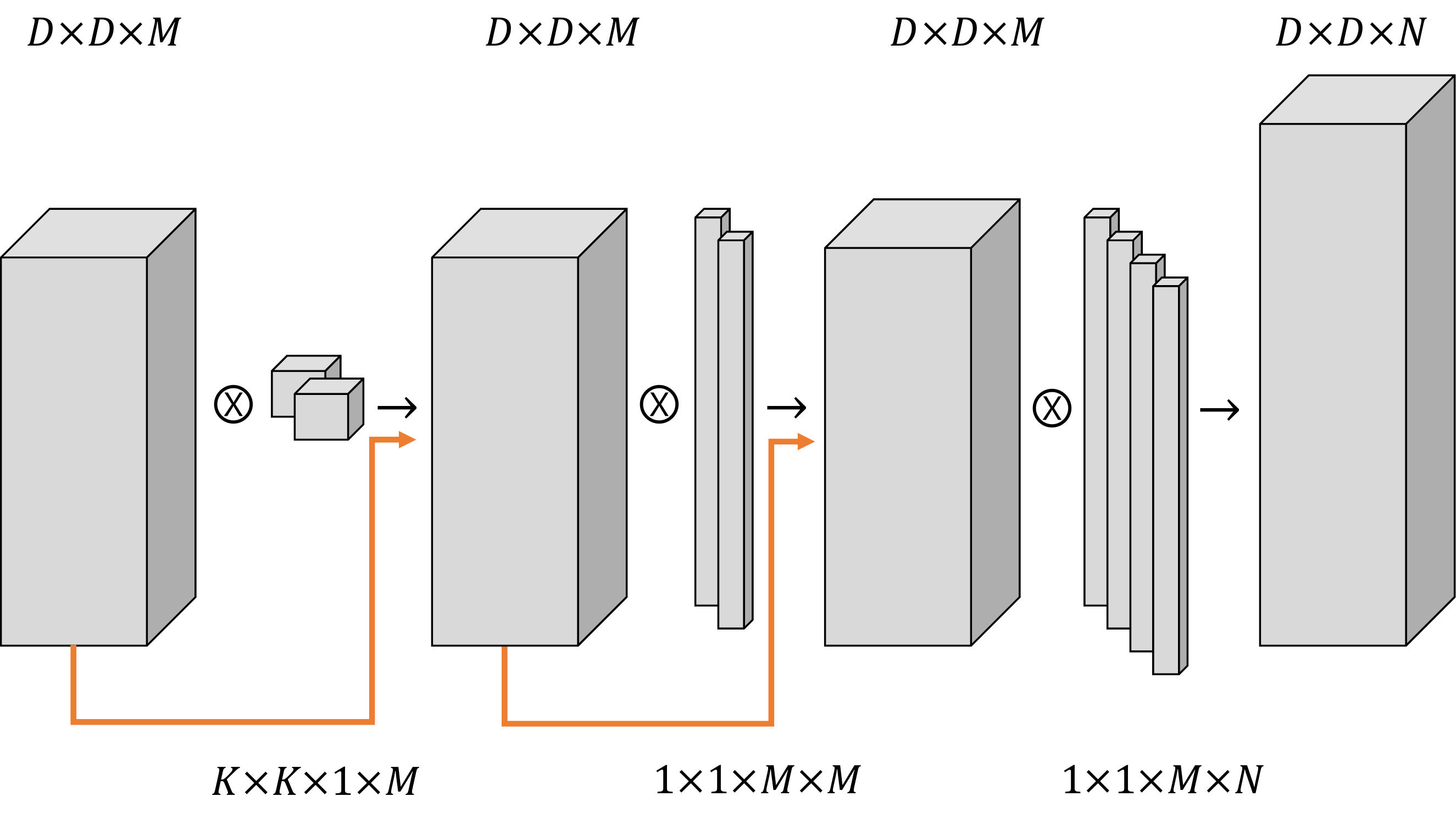}
    % \caption{demo b}
}\hspace{0.5cm}

\end{center}
    \caption{Illustration of the standard convolutional layer, depth-wise convolution in MobileNet v1 and the three MoBiNet block designs. The blocks are built with $1\times1$ binary convolution at different segments: right after the input (Pre-block), between the depth-wise and the point-wise convolution (Mid-block) and right before the output (Post-block). The designs support skip connection creations that allow avoiding loss of information and vanishing gradient when binarization.}
    \label{fig:evalpoints}
\end{figure*}

Observing that the binarization of lightweight modules in neural networks can gain considerable outcomes, in this paper we propose \textit{Mobile Binary Network} (or \textit{MoBiNet}), a network that significantly compresses MobileNet~\cite{Howard-MobileNet} to only a few megabytes while preserving good accuracy compared to other similar models~\cite{Matt-BinaryNet,CourbariauxBD15,Rastegari-xnor}. MobileNet feeds data into a depth-wise convolution, then integrates the output with a point-wise convolution, achieving an impressive classification accuracy on the ImageNet dataset~\cite{imagenet_cvpr09}. %However, the MobileNet does not completely suit low capacity devices.
%So the depth-wise layer binarization in MobileNets with comparable results to other binary neural networks can guarantee a novel improved deep compressed neural networks.
Binarizing the separable depth-wise module is not straightforward due to double-approximation issue: (i) the float-type depth-wise is an imitation of standard float-type convolution behavior and (ii) binary depth-wise is an approximation of the float-type depth-wise. Training a deep neural network consisting of binary depth-wise modules is challenging and prone to divergence due to loss of information and vanishing gradient. To tackle the problem, we introduced skip connections and channel dependency enhancement between separable layers for vanishing gradient avoidance through depth-wise layers. We propose three modules to ease the construction of skip connections: Pre-block, Mid-block, and Post-block where we can observe a convergence guarantee over training epochs. On the other hand, our MoBiNet facilitates the training from scratch, without the need of having a pre-trained model to start with. The architecture is illustrated in Fig. \ref{fig:evalpoints}. 

The main contributions of the paper are as follows:
\begin{itemize}
    \item We propose a novel compressed architecture called MoBiNet that leverages channel connections and binarizes depth-wise convolution layers. Training from scratch, the MoBiNet with binary separable convolutional layers dramatically improves speed and model size (to only a few megabytes) of neural networks while preserving comparable accuracy to state-of-the-art models. %In the section ???, we present ablation studies to prove that skip connections and our proposed blocks are very helpful to binarize depth-wise convolution layers.
    \item We propose the use of skip connections to support the non-trivial network training when binarization happens at both activations and weights of the depth-wise convolution. The MoBiNet exploits this dual binarization on the lightweight module consisting of separable convolutions.
    \item We propose the use of $K$-dependency to keep the MoBiNet small and efficient. Instead of increasing the number of convolution channels for better performance (and also larger model size), we control dependency within channels in depth-wise convolution layers.
\end{itemize}
 We experimentally show promising results of MoBiNet to solve image classification in the ImageNet dataset. The result demonstrates the potential of MoBiNet, while being efficiently small, can achieve comparable accuracy and speed as other state-of-the-art large binary neural networks~\cite{Rastegari-xnor, Liu2018BiRealNE} and even surpass them.
 
 %The paper is organized as follows. Section \ref{sec:relatedWork} introduces the current approaches for deep compressed neural networks. Section \ref{sec:MBN} presents the MoBiNet architecture. The performance evaluations are conducted in section \ref{sec:experiments}. Section \ref{sec:conclusion} concludes our work.
\section{Related works}
\label{sec:relatedWork}

\noindent\textbf{Binarized networks}. The weights in deep neural networks are often presented in full-precision or float-type values which requires much memory to store and takes time to deploy convolutional operators in the networks. There are some ideas to convert full-precision networks into binary networks. Courbariaux and Bengio~\etal~\cite{Itay-BNN,Matt-BinaryNet,CourbariauxBD15} proposed an approach to constrain weight values to either $-1$ or $1$, considered as the foundation of binary networks. To be efficient, the binary networks approximate binary values for both weights and activations that bring up comparable performance on traditional datasets: MNIST~\cite{Lecun98gradient-basedlearning}, SVHN~\cite{Yuval-nips2011}, and CIFAR-10~\cite{Krizhevsky09}. To accelerate the inference process, BinaryNet~\cite{Rastegari-xnor} used xnor and popcount operator to obtain a 7x faster runtime on GPU. However, the BinaryNet does not guarantee a good accuracy on more challenging and diverse dataset like the ImageNet. Rastegari~\etal~\cite{Rastegari-xnor} proposed Binary Weight Network (BWN) and XNOR-Net which outperform the aforementioned methods on ImageNet by more than 16\% in Top-1 accuracy. Nevertheless, BWN only requires weights to be binary while XNOR-Net binarizes the input as well.
%This approach estimates full-precision weights by solving an optimization problem which leads to multiplication between scale factors and binary weights. 
%From our observation, the BinaryNet can undermine high-level feature presentation through layers because of significant information loss when data samples have complex visual structures. In such a case,
The scale factors in XNOR-Net help retain the high-level information, even BWN on AlexNet~\cite{Alex_NIPS2012} gains 56.8\%, slightly better than its full precision (56.6\%). Another approach named Bi-Real Net~\cite{Liu2018BiRealNE} improves BNNs in terms of activations and weight binarization for ResNet~\cite{Kaiming_ResNet}, using derivative sign function approximation. It also improves weight magnitude related to the gradient, then combines with a pre-trained float-type model and gains impressive accuracy of 56.4\% and 62.2\% with ResNet-18 and ResNet-34 backbone respectively.

\noindent\textbf{Quantized networks}. The general case of binarized networks is that weight values are quantized to a fewer number of bits. Han~\etal~\cite{Han-Deep} not only pruned but also quantized unnecessary connections, then fine-tuned remaining weights with less bit representation to gain a speed-up factor. Another approach is to approximate weights as $2^k$ ~\cite{zhou2017} ($k=-2,-1,0,1$, etc.). During the estimation, the network utilizes the shift operator to make computation faster. Ternary Weight Network (TWN)~\cite{li2016ternary} is a variant of the quantized network. It defines zero as a third value for optimizing the cost of the BinaryNet. Zhu and Han~\etal~\cite{zhu2016trained} proposed an incremental approach to train TWN with ternary numbers ($\Wv_n$, 0, $\Wv_{np}$) that are learnable. These settings compress the ternary network nearly 16x in comparison with full-precision network and improve the accuracy on ImageNet of precedent ternary networks. Although quantized networks are potential, the deployment on mobile platforms is not trivial and requires much effort.

\begin{figure*}
\begin{center}
\subfigure{
    \includegraphics[scale=0.57]{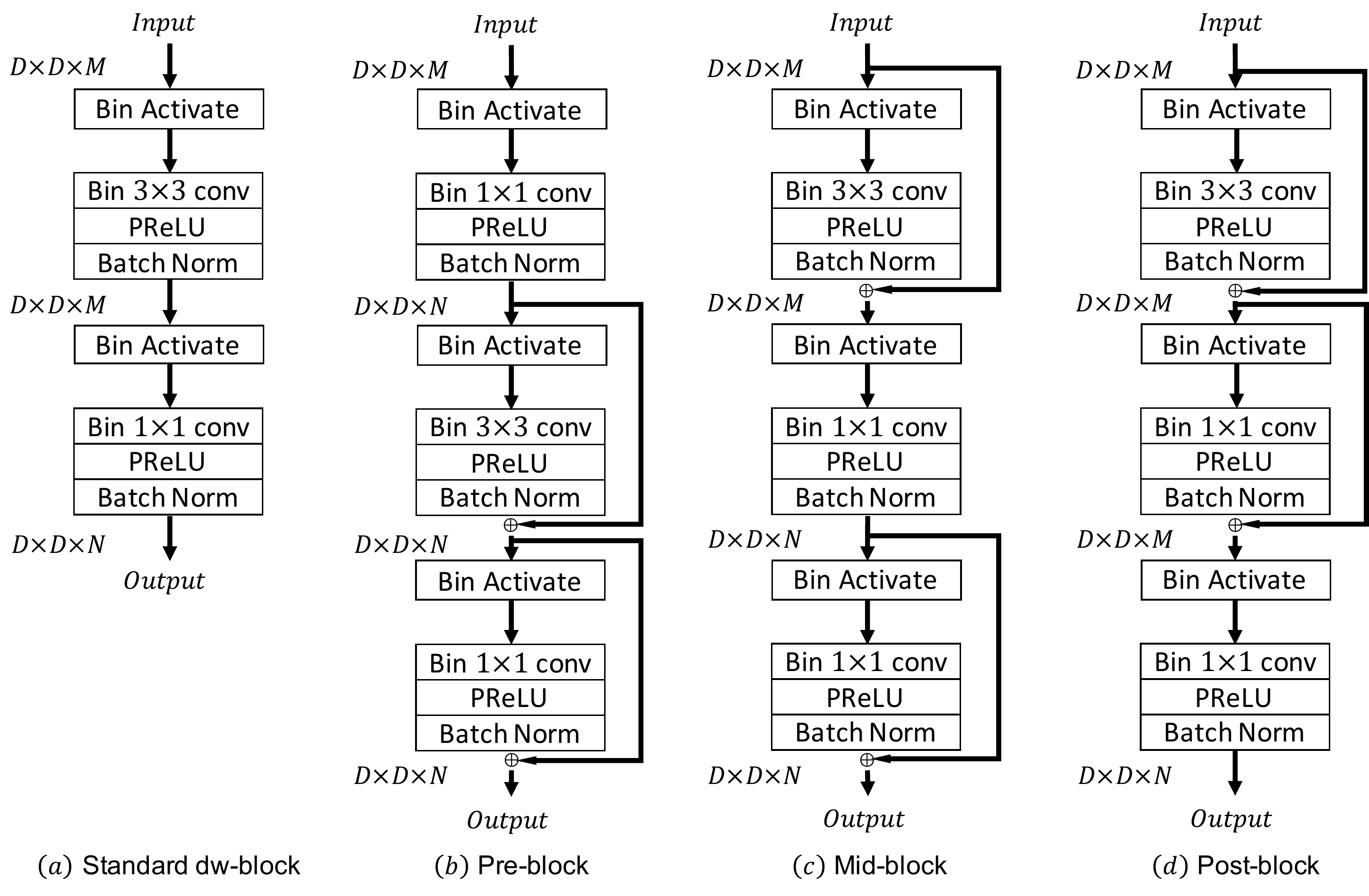}
    % \caption{demo a}
}\hspace{0.5cm}
\end{center}
    \caption{Detail of $(a)$ binary standard separable depth-wise structure in MobileNet v1 and the three binary MoBiNet block structures: $(b)$ Pre-block, $(c)$ Mid-block and $(d)$ Post-block. The binarization happens at both input and weights of every convolutional layer. The convolutions are followed by PReLU and Batch Normalization sequentially.}
    \label{fig:net-config}
\end{figure*}

\section{Mobile Binary Network - MoBiNet}
\label{sec:MBN}
In this section, we describe MoBiNet architecture, our proposed novel binary network robust for large scale image classification problem. The depth-wise convolution layers of MobileNet are not suitable to be directly binarized due to accuracy loss and training convergence issue. In MoBiNet, we modify the depth-wise convolutions to make it easier to train, work effectively with both activations and weights binarization, hence accelerate the overall network architecture efficiency. Compared to the XNOR-Net~\cite{Rastegari-xnor}, the MoBiNet generalizes the gradient update procedure adjusting the model weights through back-propagation. The scaling factors for each filter are gradually tuned in an adaptive manner with respect to the gradient update. To further boost the performance of the MoBiNet, $K$-dependency is exploited to enhance dependency within input and convolution channels. Thanks to this enhancement, the classification accuracy is significantly improved with a minor computational cost trade-off, while the model size is remarkably kept unaffected.

\subsection{MoBiNet's Architecture}
Float-type separable depth-wise convolution of MobileNet is efficiently accurate and small in size. We build MoBiNet with reuse of this module as it is core of the MobileNet v1~\cite{Howard-MobileNet} architecture. MoBiNet can be considered as a compact binary version of MobileNet. Specifically, the separable convolution layers contain two lightweight modules: a $3\times3$ depth-wise convolution with single channel-wise connection and a $1\times1$ point-wise convolution responsible for linking the depth-wise output channels. % In detail, feeding input to the depth-wise convolution layer gives separate channels as an intermediate output, and then the $1\times1$ point-wise filters merge the output channels to create a feature block of desired depth.%, gaining a new feature approximating for the feature from standard convolution layers.

However as mentioned above, the depth-wise convolution binarization can cause severe loss of information, leading to the vanishing gradient issue during training. We notice that skip connections~\cite{Kaiming_ResNet} or channels connections can reserve features through many layers, making convolution neural networks converge steadily. From this intuition to ease the skip connection creations, MoBiNet exploits an additional point-wise convolution layer at three positions: right after the input (Pre-Block), between depth-wise and point-wise convolution (Mid-Block), and right before the output (Post-Block). The purpose of the block designs is to increase skip connection operators, making separable convolution layers able to keep original features. The architecture detail is illustrated in Fig. \ref{fig:net-config}. We also replace ReLU~\cite{NairH10} with PReLU~\cite{HeZR015} which ensures the convergence stability of MoBiNet. The calculation flow at each convolution segment is as follows: Input $\rightarrow$ Binary Activation $\rightarrow$ Binary Convolution $\rightarrow$ PReLU $\rightarrow$ Batch Normalization $\rightarrow$ Output. Unlike MobileNet, to utilize skip connections MoBiNet uses Pooling layers (e.g., max/min/average pooling) to shrink spatial dimension gradually through the network. We opt average pooling for MoBiNet since it experimentally shows a better result than max and min pooling. To prove the impact of skip connections in training MoBiNet, we deploy ablation studies in section \ref{sec:ablation}.

\subsection{Train Mobile Binary Networks}
Binary networks have recently become one of the most prominent approaches in compressed deep neural networks (DNNs) designs. Binary operators can help reduce the computation cost greatly and 1-bit representation of weights supports memory saving which is crucial for mobile devices. To binarize weights and activations, most of BNNs used non-linear $sign$ function to quantize DNNs into $-1$ and $+1$ as follows:
\begin{equation} \label{eq:1}
\zv^b = \text{sign}(\av) = \left\{\begin{matrix}
 +1,& & \text{if} \ \  \av\geqslant 0, \\ 
 -1,& & \mbox{otherwise}
\end{matrix}\right.
\end{equation}
where $\zv^b$ outputs a binary value and $\av$ can be weights or activations. We follow the method described in XNOR-Net~\cite{Rastegari-xnor} to binarize weights and activations in BNNs. However, we modify the way to update parameters to make the training proper. Equation~\ref{eq:updateGradient} indicates how gradient is updated. 
%In ordinary DNNs, input data are put to each layer to obtain higher representation. , where $\wv_i\in\Rb^{k\times k \times c}$
Let's denote $\Wv =[\wv_1,\wv_2,\cdots,\wv_n]$, where $\wv_i=\left \{ \wv_{ij} \right \}_{j=1}^m$, $\wv_i\in\Rb^{k\times k \times c}$ is full-precision weight. The activation function for the forward pass of input $\xv$ is as follow:
% and $\wv^i\in \Rb^{c\times k \times k}$
\begin{equation}
\av^i = \sigma(\wv^i \xv)
\end{equation}
%
% $\Wv^b=Sign(\Wv)=[\wv_1^b,\wv_2^b,\cdots,\wv_n^b]$
% where  denotes the $i^{th}$ filter of networks at $L^{th}$ layer, and $(c,k)$ are the input channels and the kernel size respectively. To ease the mathematical denotation in the rest of this section, let's denote $\Wv^i_L$ be $\Wv$. 
%Because filters have the same size of width and height in most DNNs, we denote $k$ as a common kernel size of filters. 
Let $k\times k$ be filter kernel size, $\av^i$ is an activation output. The most expensive computation in DNNs is the dot-product of inputs and weights $\Wv\xv$. The algorithm converts both weights and activations to 1-bit representation to reduce the required amount of storage memory. It then uses a convolution without multiplication operator to facilitate and accelerate the inference stage. $\Wv$ is binarized by equation \ref{eq:1} and the optimal solution is  $\Wv^b=\text{sign}(\Wv)=\{\wv_i^b\}^n_{i=1}$, $\wv^b_i\in \{-1,+1\}^{k \times k \times c}$. Similarly, given the input $\Iv \in \Rb^{w \times h \times c}$, its binary correspondence is $\Iv^b = \text{sign}(\Iv) \in \{-1,+1\}^{w\times h \times c}$. Scaling factor vector is $\alphav \in \Rb^n$ where $\alpha_i > 0$. The approach estimates scaling factors and binarized both input and weights such that $\Wv \approx \alphav\cdot \Wv^b$ and $\Iv*\Wv \approx (\Iv^b\bigodot\Wv^b)\alphav$, where $\bigodot$ indicates xnor-bitcount operator. It can be formulated as an optimization problem:
\begin{equation} \label{eq:5}
\begin{aligned}
& \underset{\wv^b_i,\alpha_i}{\text{min}}
& & \left \| \wv_i-\alpha_i \wv^b_i  \right \|^2_2 \\
& \text{s.t}
& & \wv_i^b \in \{-1,+1\}^{k \times k \times c}, \\
&&& \alpha_i > 0 .
\end{aligned}
\end{equation}
Denote $\hat\wv_i=\left \{\alpha_i\cdot \wv_{ij} \right \}_{j=1}^m$. By taking the derivative, the optimal solution for equation \ref{eq:5} is $\wv_i^b=\text{sign}(\wv_i)$ and $\alpha_i^{*} =\frac{1}{(\wv^b_i)^{T}\wv^b_i}\left | \wv_i \right |_{l1} = \frac{1}{p}\left | \wv_i \right |_{l1}=\frac{1}{p}\sum_{j=1}^{m}\left |\wv_{ij}\right |$ where $p=k \times k \times c$. It is proved that the optimal binary weights $\wv^b$ can be directly computed using a $\textit{sign}$ function and the optimal scaling factors are the average of absolute value of $\wv_i$.  Let $\hat\Wv= \alphav\cdot\mbox{sign}(\Wv)= \{\hat\wv_i\}^n_{i=1}$ be the approximation of full-precision weights w.r.t binary weights, $\hat\wv_i=\alpha_i\cdot\mbox{sign}(\wv_i)$. The gradient of the loss function $L$ w.r.t $\wv_{ij}$ is computed as follows:
\begin{equation}
    \nabla_{\wv_{ij}}L=\left ( \frac{\partial L}{\partial \wv_{ij}} \right )^T
\end{equation}
\begin{equation}
        \frac{\partial L}{\partial \wv_{ij}}=\frac{1}{p} \mbox{sign}(\wv_{ij})\sum_{k=1}^{m}\frac{\partial L}{\partial \hat{\wv}_{ik}}\wv_{ik}^b+\frac{\partial L}{\partial \hat{\wv}_{ij}}\alpha_i\frac{\partial \wv_{ij}^b}{\partial \wv_{ij}}
\label{eq:updateGradient}
\end{equation}
In practice when training, full-precision weights are binarized in the forward pass. However in the backward pass, weights are updated with full-precision values. When the training stops, the full-precision weights are completely binarized and saved with 1-bit representation. 
\subsection{Channels Dependency Enhancement}
\begin{figure}[h!]
    \centering
    \includegraphics[scale=0.3]{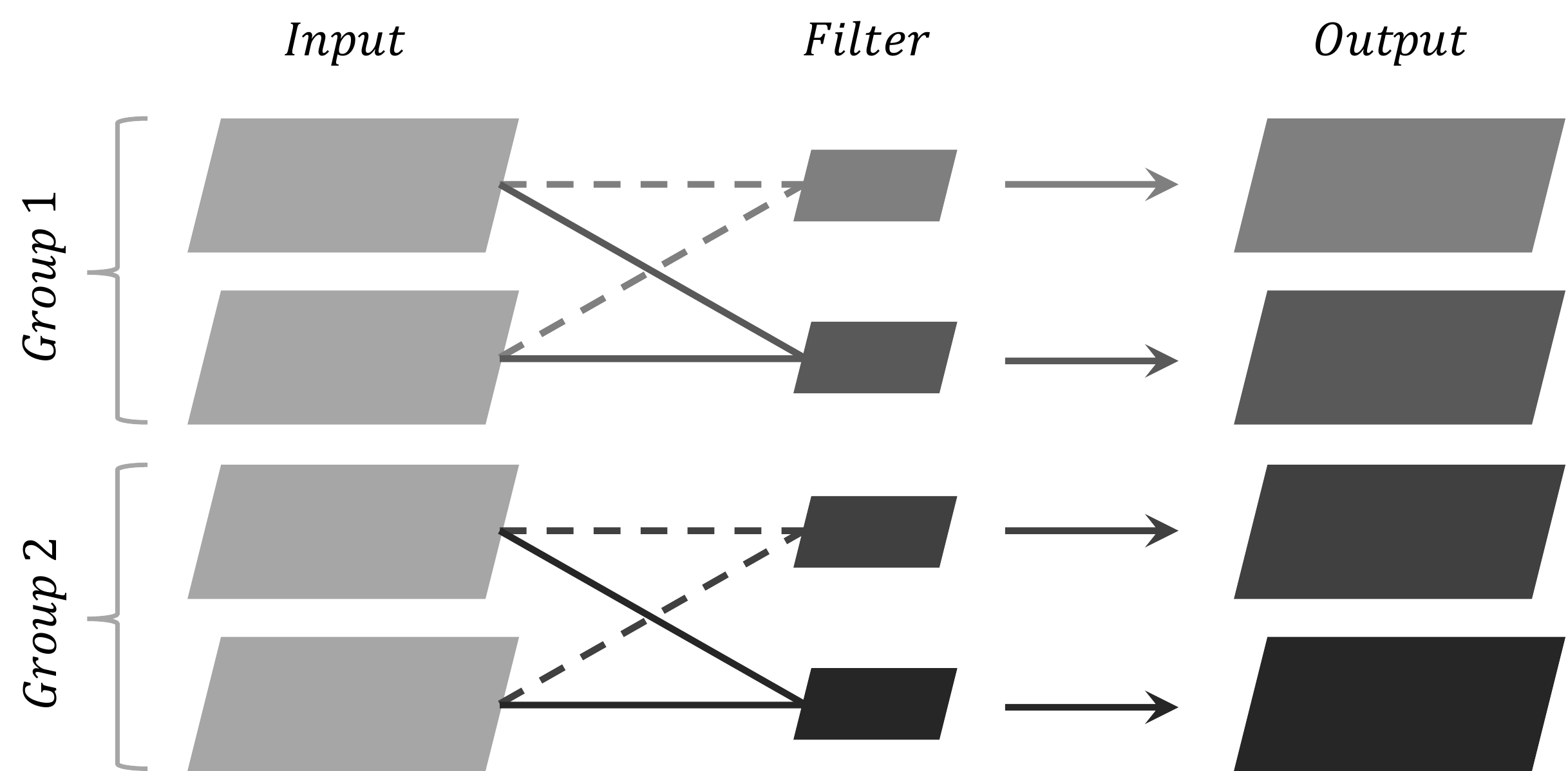}
    \caption{Channel dependency with $c=4$ and $K=1$.}
    \label{fig:dependencyChannels}
\end{figure}
In this section, we present how to improve single-layer dependency in depth-wise convolution by exploiting $K$-layer dependency, or $K$-dependency for short. The intuition is, due to the lack of channel interaction in single-layer depth-wise, the output suffers severe information loss and it is no longer efficient after being binarized. To cope with this, we provide a more complex dependency between input and convolution channels by defining the term $K$-dependency. This term allows to control the dependency level of channels and has a flavor of group convolution. Figure~\ref{fig:dependencyChannels} depicts how the $K$-dependency is set up. In the section \ref{sec:k-depen}, we explore the efficiency of the $K$-dependency by varying $K=1,2,3$ (i.e., splitting the input and convolution channels into $2^1$, $2^2$, $2^3$ groups) together with Pre-block, Mid-block and Post-block integration. The term $K$ can be also considered as group convolutions with $\#groups=\frac{c}{2^K}$, where $c$ is the number of input channels. If $K=0$, $\#groups=c$, i.e., the extreme case equivalent to depth-wise convolution. The output activation of a group $g$ can be computed by inner convolution operation within channels and filters in the same group as follows:
\begin{equation}
    \av_i^g=\sigma\left ( \sum_{j=1}^{g} \wv_i^g\cdot \xv^g_j\right )
\end{equation}
where $\av^g_i$ is the $i^{th}$ output activation at group $g$, and $\wv_i^{g}$, $\xv^g_j$ is the $i^{th}$ weight and $j^{th}$ input in group $g$. %Fig. \ref{fig:dependencyChannels} is a visualization for this representation.  
\section{Experiments and Evaluations}
\label{sec:experiments}
In this section, we analyze the efficiency our proposed MoBiNet. Three ablation studies are conducted. The first is to verify our hypothesis that skip connections are helpful for separable convolution binarization, section \ref{se:eff_skipconnect} presents comparison between mobile networks with and without skip connection. The second is to evaluate the efficiency of the three extra block design (Pre/Mid/Post-block) added in MoBiNet and compare with the binary original MobileNet. The third is, in section \ref{sec:prelu} we present efficiency of replacing ReLU with PReLU for MoBiNet to increase accuracy. In addition, in \ref{sec:k-depen} we also demonstrate the impact of $K$-dependency to the enhancement of MoBiNet, under several selections of hyper-parameter $K$ ($K=0,1,2,3$). Finally, to indicate that our MoBiNet can achieve comparable outcomes with the other binary neural networks we analyze its result with the state-of-the-art binary neural networks on five metrics: Top-1, Top-5 classification accuracy, number of parameters, and number of FLOPs. Regarding the later two metrics, we also point out numerical gain factors that MoBiNet achieves.
\subsection{Dataset and Implementation Details}
\label{sec:data_impl}
\noindent
\textbf{Dataset}. %We evaluate the performance of MoBiNet on the challenge task of visual image classification. 
The image classification training and evaluation are carried on ILSVRC12 ImageNet dataset. The ImageNet is a large scale and diverse dataset containing 1.2M images for training and 50K for testing. They are classified into 1000 categories such as lionfish, airliner, red panda, etc. We opt to use ImageNet because it is such popular that the XNOR-Net~\cite{Rastegari-xnor} and the Bi-RealNet also use for benchmarking purpose, thus making the comparison with MoBiNet fair. Each image in ImageNet is scaled to the size of $256\times256$ pixels. For training, images are randomly cropped to fit $224\times224$. For testing, we apply center crop with the same size of $224\times224$.\\

\noindent \textbf{Implementation details}. All experiments are implemented using PyTorch deep learning framework~\cite{paszke2017automatic}. We run tests with Pre-block, Mid-block, and Post-block designs for MoBiNet. During the training of binary neural networks, we process the float-type model and the binary model concurrently. That means the floating-point weights are stored in RAM for weight update through backpropagation and they are binarized to compute activation in the next layer. For every experiment, we train the model in $50$ epochs and learning rates are set to $10^{-3}$ to train the Pre-block and $10^{-4}$ for training the Mid-block and Post-block. We choose the mini batch size $128$, weight decay $10^{-5}$ and momentum $0.9$ with Adam optimizer~\cite{kingma2014method}. At some points, the convergence becomes slow and requires manual learning rate update. To foster the training, the three blocks receive the same multiplicative decrease factor $0.1$ at different points of time: the Mid-block is at $20^{th}$ and $30^{th}$ epoch whereas the Pre-block and the Post-block are at $15^{th}$ and $45^{th}$ epoch.

%The learning rate starts updating by dividing for $10$ at $20^{th}$ and $15^{th}$ epoch for mid-block and pre-block, post-block respectively. It continue updating at $30^{th}$ and $45^{th}$ epoch later. 

In the first convolution layer, feature representation often has a low depth dimension (e.g., 3 for visual RGB image). To avoid the loss of crucial information, the first convolution layer is not binarized. Also, the output features extracted by MoBiNet are not binarized either to increase the capability of classification. These make the binary model compression factor slightly less than 32x in the context of 32-bit float type and binary type.
% Like XNOR-Net~\cite{Rastegari-xnor}, the first convolution and the classifier layer should not be binarized to reserve information. 
For the rest of the network, both input and weights are binarized, then multiplied with the according scaling factors $\alphav_i$. Weights are updated following Equation~\ref{eq:updateGradient}. When the training finished, the models are saved with 1-bit representation to reduce memory storage for inference.
\subsection{Ablation studies}
\label{sec:ablation}
\subsubsection{The Efficiency of Skip Connections In MoBiNet}
\label{se:eff_skipconnect}
\begin{figure}[t]
\hspace*{-0.3cm}
    \centering
    \includegraphics[scale=0.42]{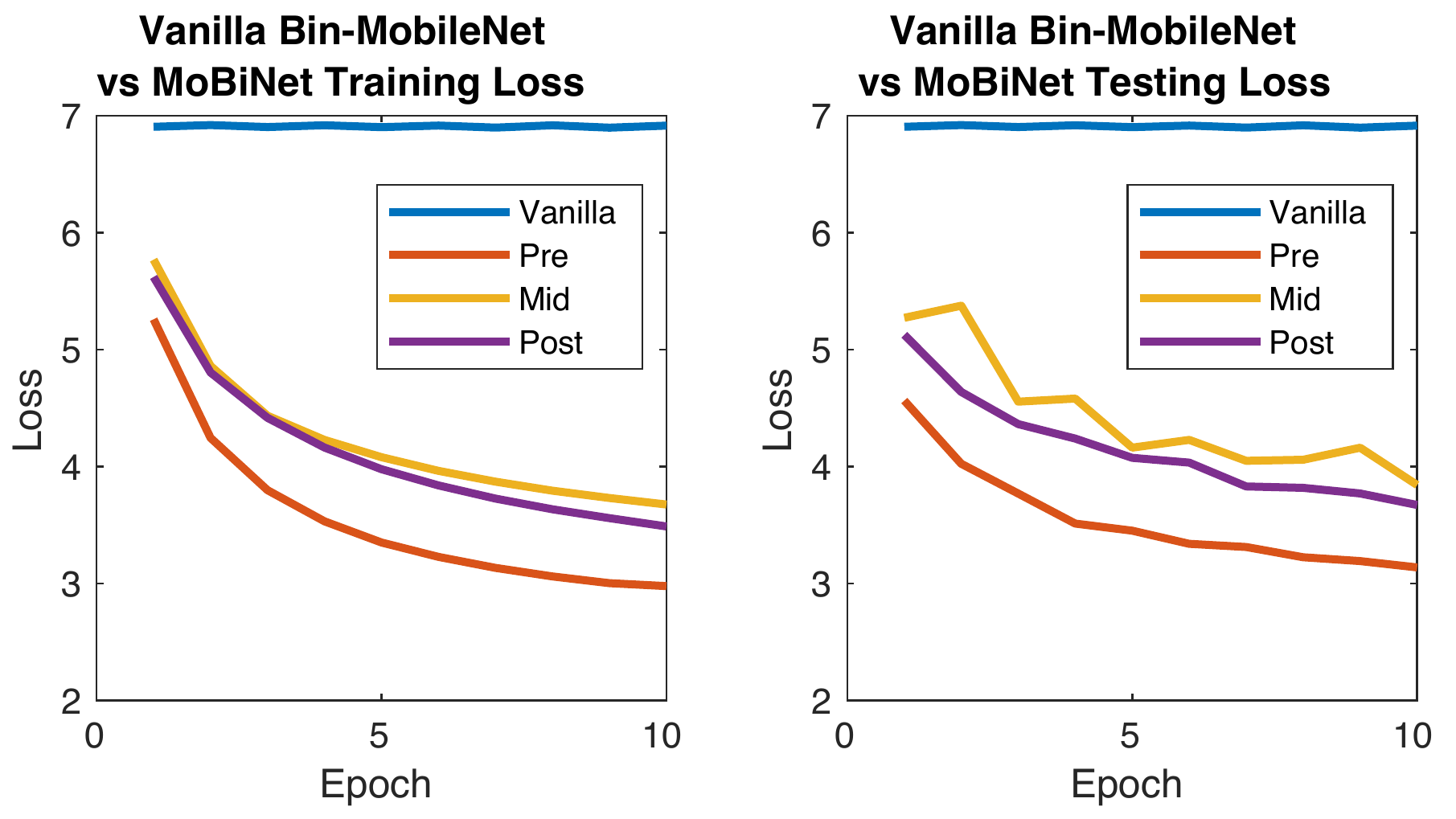}
    % \caption{Loss of training and testing the binary vanilla MobileNet and our MoBiNet.}
    \caption{Convergence of Vanilla Bin-MobileNet vs MoBiNet.}
    \label{fig:vanilla}
\end{figure}

Recall that training binary separable convolution is prone to unstable convergence due to considerable loss of information through layers. The fact that training loss diverges indicates a serious issue in the mobile binary neural network. To overcome this obstacle, we deploy skip connections between separable convolution layers to aid the training. In this experiment, we test the efficiency of the skip connections in MoBiNet. The skip connection performance of Pre-block, Mid-block and Post-block is reported for training and test phase. We also present the loss values of without-skip-connection in vanilla binary Mobilenet through epoches in Fig. \ref{fig:vanilla} to show that training does not converge.
%which intuitively give an extremely bad accuracy. 
\begin{figure}[h]
%\hspace*{-0.5cm}   
\centering
    \includegraphics[scale=0.43]{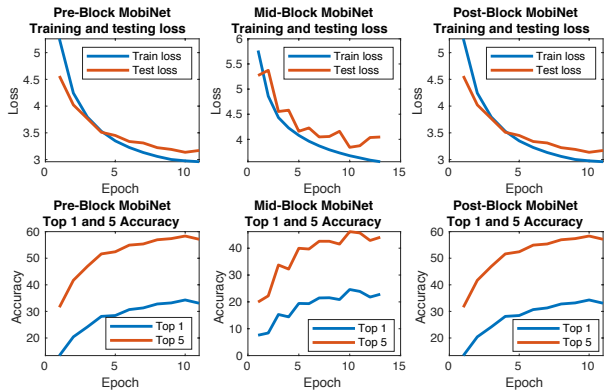}
    \caption{Training and testing performance (with the loss (top) and accuracy (bottom)) when using skip connections in the three block designs.}
    \label{fig:skip}
\end{figure}
The performance of training with skip connections is shown in Figure~\ref{fig:skip} with the three blocks. We pick $K=0$, i.e., zero-dependency and equivalent to the original depth-wise convolution. We stopped the training at $10^{th}$ and $15^{th}$ epoch to see the effect. Even though the convergence slope and accuracy are not decent due to the zero-dependency, one can observe that the training goes steady over time.
%The models converge with a fast pace and the accuracy also increases over time.
It confirms that our hypothesis of using skip connections for separable convolution binarization can deal with the divergence issue by limiting the loss of information. Without loss of generality, it is applicable to binarize weights of not only in MoBiNet but other neural networks~\cite{mehta2018espnet, Zhang2018ShuffleNetAE,Ma_2018_ECCV,Fran-Xcep} using depth-wise convolution as well. 
\subsubsection{Efficiency of MoBiNet modules: Pre-block, Mid-block and Post-block}
Skip connections clearly help mobile neural network with separable convolution layers converge and achieve reasonable results. To further increase the efficiency of the skip connection and explore its potential, we proposed an incremental $1\times1$ point-wise convolution layer to help up\-sample the features and support the skip connection creation. %at positions of depth-wise convolution layers, removing the issue of significant loss information.
The $1\times1$ point-wise convolution layer is placed at three different positions: right after the input (Pre-block), between depth-wise and point-wise convolution (Mid-block), and right before the output  (Post-block). %They are pre-block, mid-block, and post-block. Pre-block, mid-block and post-block adds a point-wise layer at the beginning, middle, and the end of depth-wise convolution layers.
Figure~\ref{fig:net-config} illustrates the locations of the $1\times1$ convolution. In this evaluation, we pick the Pre-block design as a representative to conduct experiments in two scenarios: with and without the block design, then analyze the impact. The authors also observe a consistent result in the case of Mid-block and Post-block.

The experiments are set up as follows:
\begin{itemize}
    \item We construct MoBiNet with and without Pre-block with zero-dependency depth-wise convolution layers (i.e., choosing $K=0$).
    \item The MoBiNet is built with aforementioned hyper-parameters as in section \ref{sec:data_impl}. 
    % {\color{red}Section...}
    \item When training MoBiNet with and without Pre-block design, we run in 20 epochs, starting with an initial learning rate of $10^{-4}$ and updating the rate at the $15^{th}$ epoch with a decay factor $0.1$. The input and weights are both binarized except the first convolutional layer and the classifier layer right before the classification output. 
    \item We report Top-1 and Top-5 accuracy for comparison, then plot the training curve through epochs.
\end{itemize}
\begin{table}
\centering
\begin{tabular}{ |c|c|c|c| } 
\hline
Networks & Top-1 Accuracy (\%) & Top-5 Accuracy (\%) \\
\hline
% \multirow{3}{4em}{Multiple row} & cell2 & cell3 \\ 
% w/o blocks & 33.31 & 57.30 \\ 
W/o blocks & 25.46 & 47.34 \\ 
Pre-block & $\textbf{35.86}$ & $\textbf{59.46}$ \\ 
% Mid-block & $\textbf{34.23}$ & $\textbf{57.51}$ \\ 
% Post-block & 32.74 & 56.17 \\ 
\hline
\end{tabular}
\bigskip
\caption{Top-1 and Top-5 accuracy (in percentage) comparison between MoBiNet with and without the Pre-block design}
\label{tab:blocks}
\end{table}

\begin{figure}
% \hspace*{-0.0cm}   
    \centering
    \includegraphics[scale=0.34]{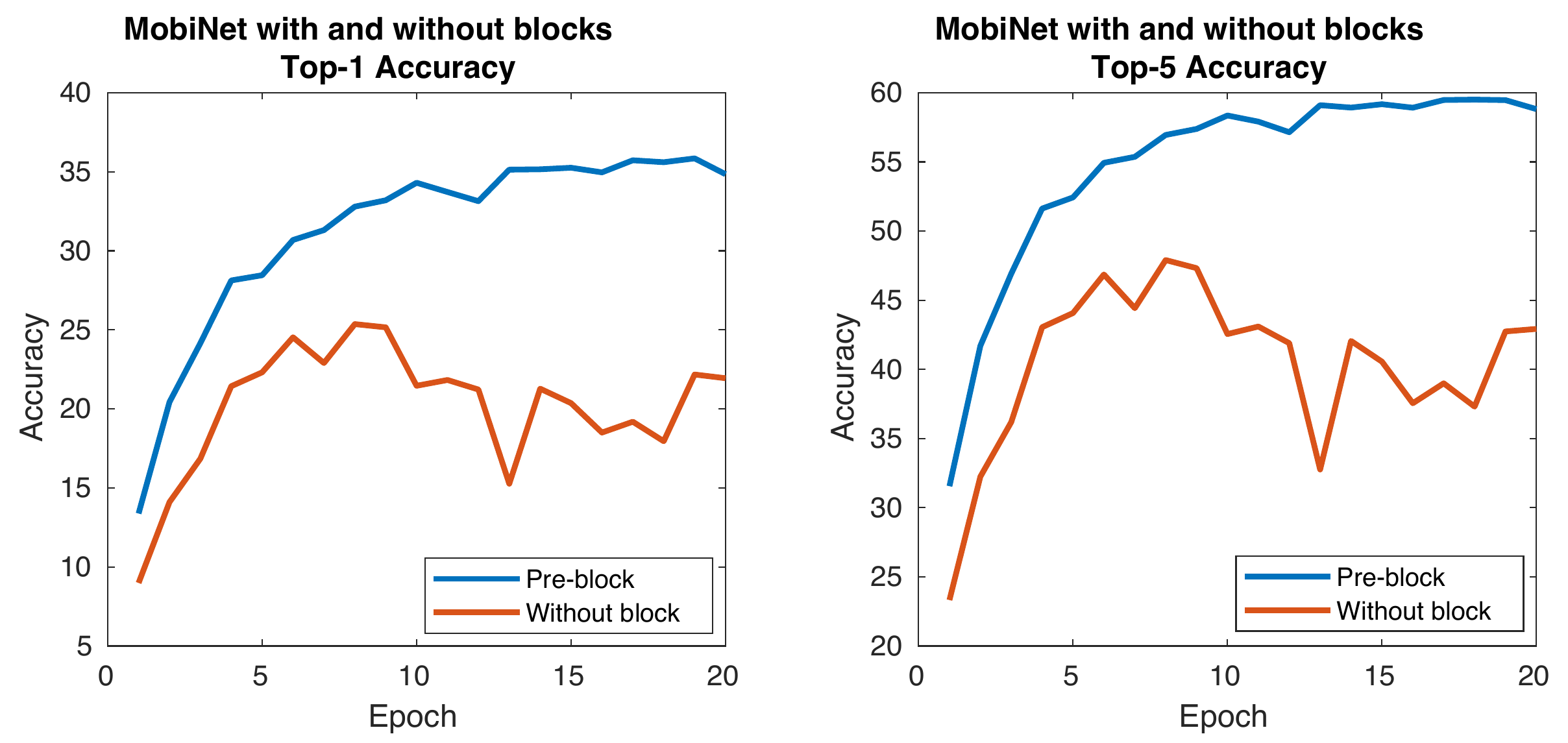}
    \caption{Training performance: Top-1 and Top-5 accuracy of MoBiNet with and without Pre-block through $20$ first epochs.}
    \label{fig:testblocks}
\end{figure}

The results are shown in Table \ref{tab:blocks} and training performance is in Figure~\ref{fig:testblocks}. Convergence of MoBiNet with the blocks is more stable and gains better accuracy while the without-block design fluctuates as the training goes on. It confirms that an extra $1\times1$ point-wise layer for MoBiNet's block design can ease the training of the separable convolution layer.
\subsubsection{Efficiency of PReLU layer used in MoBiNet}
\label{sec:prelu}
To further ameliorate the MoBiNet, we replace ReLU with PReLU as it experimentally shows a better feature representation throughout the binary network. In addition, PReLU layers help MoBiNet increase accuracy without incurring computational cost or parameter storage. The efficiency of PReLU is presented in Table  \ref{tab:relu}. We trained the Mid-block designe with $K=0,1,2,3$ where settings of ReLu and PReLU are exactly the same. 

\begin{table}
\centering
\begin{tabular}{ |c|c|c|c| } 
\hline
Networks & ReLU & PReLU \\
\hline
% Mid-block ($K=0$)& 32.44 / 55.78 & \textbf{34.73 / 58.28} \\ 
% Mid-block ($K=1$)& 48.52 / 72.45 & \textbf{50.41 / 73.98} \\
% Mid-block ($K=2$)& 49.60 / 73.41 & \textbf{52.25 / 75.32} \\
% Mid-block ($K=3$)& 51.36 / 75.14 & \textbf{53.47 / 76.46} \\
Mid-block ($K=0$)& 32.44 & \textbf{34.73} \\ 
Mid-block ($K=1$)& 48.52 & \textbf{50.41} \\
Mid-block ($K=2$)& 49.60 & \textbf{52.25} \\
Mid-block ($K=3$)& 51.36& \textbf{53.47} \\
\hline
\end{tabular}
\bigskip
\caption{Benefit of using PReLU instead of ReLU. The accuracy (in percentage) is reported with Top-1 accuracy on ImageNet. The use of PReLU for MoBiNet improves the accuracy approximately $2\%$ in comparison with ReLU.}
\label{tab:relu}
\end{table}

\subsection{$K$-Dependency for MoBiNet Enhancement}
\label{sec:k-depen}
\begin{table*}[h!]
\hspace*{-0.5cm} 
\begin{center}
\begin{tabular}{|l|c|c|c|c|c|c|}
\hline
\specialcell{Networks} & K & \specialcell{Top-1 \\ accuracy(\%) }& \specialcell{Top-5 \\ accuracy(\%)}&\specialcell{Mem \\usage\\(MB)} &\specialcell{FLOPs (M)}&\specialcell{Speed \\ up} \\
\hline\hline
\multirow{4}{0em}{Pre-block} & 0 & 35.86 & 59.46 &\multirow{4}{2em}{4.60}&45.87&$12.40\times$\\
 & 1 & 48.93 & 72.83&&46.57&$12.23\times$\\
 & 2 & 50.74 & 74.22&&47.97&$11.86\times$\\
 & 3 & 51.41 & 74.75&&50.76&$11.21\times$\\
\hline\hline
\multirow{4}{0em}{Mid-block} & 0 & 34.73 & 58.28&\multirow{4}{2em}{4.60}& 45.61&$12.40\times$\\
 & 1 & 50.41 & 73.98&&46.04&$12.48\times$\\
 & 2 & 52.25 & 75.32&&46.90&$12.36\times$\\
 & 3 & \textbf{53.47} & \textbf{76.46}&&48.62&$12.13\times$\\
\hline\hline
\multirow{4}{0em}{Post-block} & 0 & 33.58 & 56.69 &\multirow{4}{2em}{4.50}&35.37&$16.09\times$\\
 & 1 & 46.83 & 70.57&&35.80&$15.89\times$\\
 & 2 & 48.22 & 71.97&&36.67&$15.52\times$\\
 & 3 & 49.88& 73.27&&38.39&$14.82\times$\\
 \hline\hline
 \specialcell{Mobile\\Net~\cite{Howard-MobileNet}}  & Full & 70.9 & 89.90 &32.40&569&$1.00\times$\\
 \hline
\end{tabular}
\end{center}
\bigskip
\caption{The $K$-dependency performance of MoBiNet including Top-1 and Top-5 accuracy in percentage, memory usage in Megabytes (Pytorch model saving~\cite{paszke2017automatic}), and FLOPs computed as in ~\cite{Liu2018BiRealNE}. The comparison with full-precision MobileNet regards accuracy and inference speed-up factor. Among the three block designs, the Mid-block yields better accuracy than Pre-block and Post-block. This is because the Mid-block exploits skip connections to preserve both original and scaled features, unlike Pre-block (only scaled feature) and Post-block (only original feature).}
\label{ta:k-depend}
\end{table*}
\begin{table*}[h!]
\hspace*{-0.5cm} 
\begin{center}
\begin{tabular}{|l|c|c|c|c|c|c|}
\hline
\specialcell{Networks} & \specialcell{Top-1\\accuracy(\%)}&\specialcell{Top-5\\accuracy(\%)}&\specialcell{ \#Params\\(M)}&Saving&FLOPs&\specialcell{Speed\\up}\\
\hline\hline
Binary Connect~\cite{CourbariauxBD15}& 35.40 & 61.00&-&-&-&-\\
BNNs~\cite{NIPS2016_6573_BNN}& 42.20 & 67.10&-&-&-&-\\
ABC-Net~\cite{NIPS2017_6638_TowardsAcc}&42.70&67.60&-&-&-&-\\
DoReFa-Net~\cite{zhou2016dorefa}&43.60&-&-&-&-&-\\
BWN~\cite{Dong2019}&39.20&64.70&-&-&-&-\\
SQ-BWN~\cite{Dong2019}&45.50&70.60.07&-&-&-&-\\
XNOR-AlexNet~\cite{Rastegari-xnor}& 44.20& 69.20&62.38&$0.19\times$&$1.38\times10^8$&$1.18\times$\\
XNOR-ResNet-18~\cite{Rastegari-xnor}& 51.20& 69.30&11.7&$1.00\times$&$1.67\times10^8$&$0.97\times$\\
% xnor-ResNet-34\cite{Rastegari-xnor}& -& -&21.8&$2.56\times$&$1.98\times10^8$&$1.21\times$\\
Bi-RealNet-18~\cite{Liu2018BiRealNE}&56.40&79.50&11.70&$1.00\times$&$1.63\times10^8$&$1.00\times$\\
% Bi-RealNet-34~\cite{Liu2018BiRealNE}&62.20&83.90&21.80&$0.54\times$&$1.93\times10^8$&$0.84\times$\\
\hline\hline
% MoBiNet-Mid (K=1) &\textbf{50.41}&\textbf{73.98}&&&&\\
% MoBiNet-Mid (K=2) &\textbf{52.25}&\textbf{75.32}&&&&\\
\specialcell{MoBiNet-Mid ($K=3$)} &\textbf{53.47}&\textbf{76.46}&\textbf{8.50}&$\textbf{1.38}\times$&$\textbf{0.49}\times\textbf{10}^8$&$\textbf{3.33}\times$\\
\specialcell{MoBiNet-Mid ($K=4$)} &\textbf{54.40}&\textbf{77.50}&\textbf{8.75}&$\textbf{1.34}\times$&$\textbf{0.52}\times\textbf{10}^8$&$\textbf{3.13}\times$\\
\hline
\end{tabular}
\end{center}
\bigskip
\caption{The Top-1 and Top-5 accuracy comparison between MoBiNet Mid-Block and the state-of-the-art. The Bi-RealNet-18 is selected as a baseline model to make comparisons.}
\label{ta:com_acc}
\end{table*}
The separable convolution layers are limited for feature representation because there is no interaction between layers. To improve this capability for MoBiNet, we proposed a novel method using $K$-dependency to augment correlation within separate channels. In this section, we show that the factor $K$ plays a key role to boost the accuracy of MoBiNet. The larger the $K$, the deeper the channels correlate and hence the more adequately the feature is represented after being convolved. Our selections for the level of dependency are $K=0,1,2,3$. We report Top-1 and Top-5 accuracy on ImageNet data. 
% For memory usage, we compute parameters with Mega bits: $\#\mbox{float-parameters} \times 32 \ \mbox{bits}  + 1 \ \mbox{bits} \times \#\mbox{binary-parameters}$. 
The number of FLOPs is computed in a similar way mentioned in ~\cite{Liu2018BiRealNE} to ensure the comparison fairness with the other binary networks. The results are reported in Table~\ref{ta:k-depend}. Generally speaking, the MoBiNet speeds up the inference stage by $11\times$ compared to MobileNet~\cite{Howard-MobileNet}. The best performance we can achieve is with Mid-block design: Top-1 accuracy of \textbf{53.47}\% with $K=3$ and speed-up factor $\textbf{12}\times$. %while we only drop $17.43\%$ Top-1 accuracy.
% \subsection{Comparison with the state-of-the-art}
% \label{sec:comparison}
%To prove that our proposed MoBiNet gains the comparable results, in this experiment we compare
%The results in table \ref{ta:k-depend} show that the mid-block module yields the better accuracy. This is because Mid-block exploits skip connections to preserve original and scaled feature, unlike Pre-block (only scaled feature) and Post-block (only original feature).
In Table~\ref{ta:com_acc}, we show the efficiency of MoBiNet against several recent models: XNOR-Net~\cite{Rastegari-xnor}, Bi-RealNet~\cite{Liu2018BiRealNE}, ABC-Net~\cite{NIPS2017_6638_TowardsAcc}, DoReFa-Net~\cite{zhou2016dorefa}, SQ-BWN~\cite{Dong2019}, etc.
%and Binary Neural Networks~\cite{Matt-BinaryNet} achieve impressive outcomes for challenge visual task of image classification with both weights and activations binarization.
%presents a comparison between our MoBiNet and these networks. We report the performance including the accuracy and efficiency of networks.
For efficiency comparison, we opt for Bi-RealNet-18 as a baseline because this approach is the most recent and efficient binary approach. Our MoBiNet, with Mid-block design and $K=4$, achieves a speed-up factor of $\textbf{3.13}\times$ and a model saving factor of $\textbf{1.34}\times$. Furthermore, in terms of accuracy  MoBiNet outperforms all of the precedent methods except Bi-RealNet that we are $2\%$ behind. However on the ImageNet dataset, to reproduce the accuracy of the Bi-RealNet, a pretrained model is required whereas the MoBiNet has the freedom to start the training from scratch. Furthermore, one can bridge the accuracy gap by increasing the factor $K>4$ with an acceptable FLOPs trade-off.  %Although Bi-RealNet gets impressive accuracy, the model has some limitations. The architecture still keeps branched convolution layers in full-precision (the branched convolution layers make input the same dimension for skip connection). Bi-RealNet needs pre-trained models to gain considerable outcomes while it takes a long time for these models to converge on very large scale data of ImageNet. Nevertheless, Our MoBiNet binarizes both input and weights at all layers except to first convolution and the classifier layers in an already-compressed network's structure of MobileNet. All our models are trained on ImageNet from scratch with an ordinary training procedure. 

% \begin{table}
% \hspace*{-0.5cm} 
% \begin{center}
% \begin{tabular}{|l|c|c|c|}
% \hline
% \specialcell{Networks} & \specialcell{Mem. \\ usage  \\ (Mb)}& FLOPs&\specialcell{Speed \\ up}\\
% \hline\hline
% % Binary Connect~\cite{CourbariauxBD15}& 35.40 & 61.00\\
% % BNNs~\cite{NIPS2016_6573_BNN}& 42.20 & 67.10\\
% % xnor-AlexNet\cite{Rastegari-xnor}& 44.20& 69.20\\
% % xnor-ResNet-18\cite{Liu2018BiRealNE}& 51.20& 69.30\\
% % ABC-Net~\cite{NIPS2017_6638_TowardsAcc}&42.70&67.60\\
% Bi-RealNet-18~\cite{Liu2018BiRealNE}&33.6&$1.63\times10^8$&$3.35\times$\\
% XNOR-Net-ResNet18 &33.7&$1.67\times10^8$&$3.44\times$\\
% ResNet18-Full &374.10&$1.81\times10^9$&$37.24\times$\\
% \hline
% Bi-RealNet-34~\cite{Liu2018BiRealNE}&43.70&$1.93\times10^8$&$3.97\times$\\
% XNOR-Net-ResNet34 &43.90&$1.98\times10^8$&$4.07\times$\\
% ResNet34-Full &679.30&$3.66\times10^9$&$75.18\times$\\
% \hline\hline
% MoBiNet-Mid (K=3) &-&$\textbf{4.86}\times\textbf{10}^7$&$1.00\times$\\
% \hline
% \end{tabular}
% \end{center}
% \bigskip
% \caption{The memory usage and speed comparison between MoBiNet and the state-of-the-art.}
% \label{ta:com_mem}
% \end{table}
\section{Conclusion}
\label{sec:conclusion}
We introduced MoBiNet, the very first work for lightweight module binarization. We explore the training efficiency of the binarization with the support of skip connection, the three block designs and the $K$-dependency. The construction of MoBiNet facilitates the non-trivial training of a binary network, and without loss of generality the MoBiNet can be used for tasks other than image classification, for instance feature extraction, object detection, etc. With the intuitive design of MoBiNet, we can go further in the future and shed the light on novel solutions for deep compressed lightweight neural network problem.
% \newpage

{\small
\bibliographystyle{ieee}
\bibliography{egpaper_final}
}

\end{document}